\documentclass[10pt,twocolumn,letterpaper]{article}

\usepackage{cvpr}              
\usepackage{array}
\usepackage{multicol}
\usepackage{multirow}
\usepackage{pifont}
\usepackage{tabularx}
\usepackage{indentfirst}
\usepackage{makecell}
\usepackage{fancyhdr}
\usepackage{xcolor}
\usepackage{diagbox}

\definecolor{cvprblue}{rgb}{0.21,0.49,0.74}
\usepackage[pagebackref,breaklinks,colorlinks,allcolors=cvprblue]{hyperref}

\def\paperID{00783} 
\def\confName{CVPR}
\def\confYear{2025}
\definecolor{my_url}{RGB}{251,52,156}
\title{FASTer: Focal Token Acquiring-and-Scaling Transformer 

for Long-term 3D Object Detection}

\author{Chenxu Dang \hspace{0.6em} Zaipeng Duan \hspace{0.6em} PeiAn\hspace{0.6em}  Xinmin Zhang \hspace{0.6em} Xuzhong Hu \hspace{0.6em} Jie Ma\textsuperscript{*}\\
Huazhong University of Science and Technology\\}

\renewcommand{\paragraph}[1]{%
    \par\vspace{1.2ex}
    \par\noindent
    \textbf{#1}\hspace{0.5em}%
}

\begin{document}

\maketitle
\begin{abstract}

Recent top-performing temporal 3D detectors based on Lidars have increasingly adopted region-based paradigms. They first generate coarse proposals, followed by encoding and fusing regional features. However, indiscriminate sampling and fusion often overlook the varying contributions of individual points and lead to exponentially increased complexity as the number of input frames grows. Moreover, arbitrary result-level concatenation limits the global information extraction. In this paper, we propose a \textbf{F}ocal Token \textbf{A}cquring-and-\textbf{S}caling \textbf{T}ransform\textbf{er} (FASTer), which dynamically selects focal tokens and condenses token sequences in an adaptive and lightweight manner. Emphasizing the contribution of individual tokens, we propose a simple but effective Adaptive Scaling mechanism to capture geometric contexts while sifting out focal points. Adaptively storing and processing only focal points in historical frames dramatically reduces the overall complexity. Furthermore, a novel Grouped Hierarchical Fusion strategy is proposed, progressively performing sequence scaling and Intra-Group Fusion operations to facilitate the exchange of global spatial and temporal information. Experiments on the Waymo Open Dataset demonstrate that our FASTer significantly outperforms other state-of-the-art detectors in both performance and efficiency while also exhibiting improved flexibility and robustness. 
The code is available at \url{https://github.com/MSunDYY/FASTer.git}.
\end{abstract}    
\section{Introduction}
\thispagestyle{plain}
\label{sec:intro}

Lidar-based 3D object detection has received extensive attention from academia and industry in the past few years, owing to its ability to accurately percept the environment, which is critical in autonomous driving applications. 

\begin{figure}
  \centering
   \includegraphics[width=1.0\linewidth]{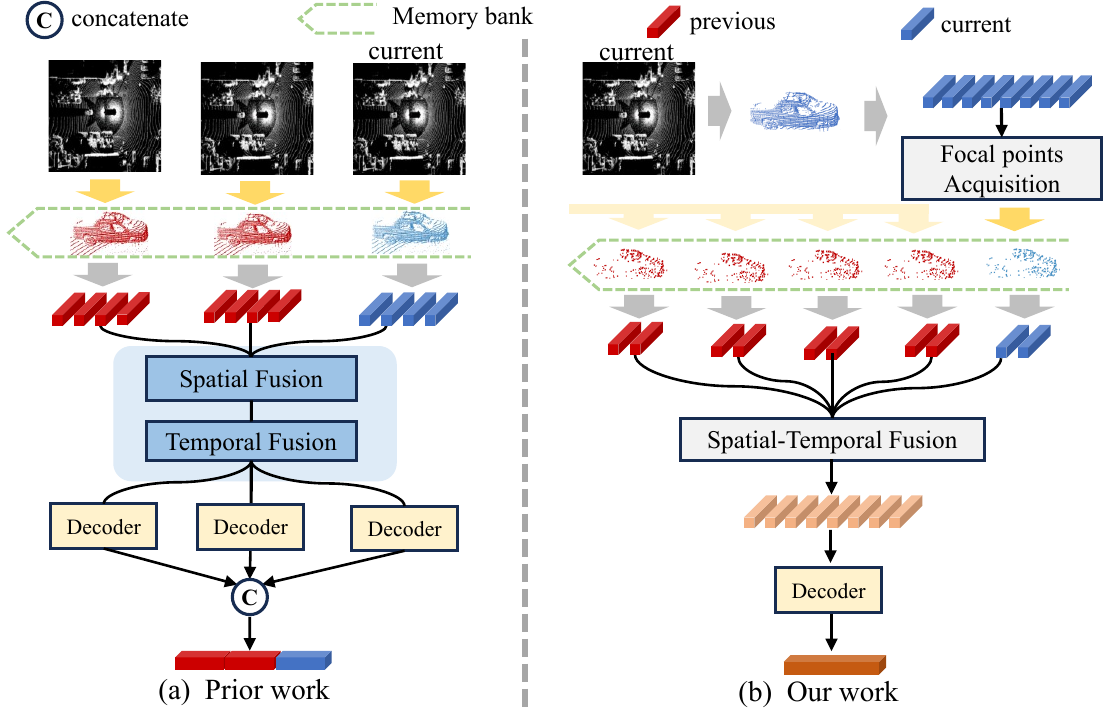}
   \caption{Comparison of different temporal-based schemas. (a) Previous methods store and collect points from the complete historical point cloud, and concatenate them on the channels after sequential spatial and temporal fusions. (b) Our method dynamically obtains, stores and encodes notable points, condenses the longer sequences into one long sequence through the integrated Spatial-Temporal Fusion module.
   }
   \label{fig1}
\end{figure}

However, the inherent sparsity of point clouds limits the performance of conventional detectors. Therefore, researchers have sought to exploit temporal information from point cloud sequences. A common approach is to concatenate multiple frames of points\cite{centerpoint} or features\cite{faf,mgtanet}, but this works within short clips. 



To full leverage longer lidar sequences, many two-stage approaches initially generate coarse proposals (e.g., boxes\cite{3dman,pvrcnn,voxelrcnn,pointconv} or queries\cite{centerformer,transfusion}). Some of them \cite{voxelrcnn,pvrcnn} fuse information at the feature map level, which, however, loses the fine-grained details in points. Recently, a series of region-based methods\cite{ct3d,mppnet,msf,ptt} have gained popularity, which abandon feature maps rather to sample points directly from the region of interest, followed by point-box embedding and regional fusion, achieving superior performance.
  
Earlier works\cite{msf,mppnet} follow an encode-fuse-concatenate routine, as shown in \cref{fig1}(a). They alternate between spatial and temporal fusion, concatenating the decoding outputs of all frames along the channel dimension. However, as the number of frames grows, their storage and computational demands increase exponentially, limiting the model's capacity of handling longer sequences. Moreover, the simple result-level concatenation also struggles to capture and exchange global information. Recently, PTT\cite{ptt} introduced a pipeline that models current points and historical box trajectories, but this sacrifices the semantics in historical points and heavily relies on the recall rate of proposals to ensure the box-level tracking.

In order to solve the above problems, in this paper, we propose a \textbf{F}ocal Token \textbf{A}cquiring-and-\textbf{S}caling \textbf{T}ransform\textbf{er} dubbed FASTer. We observe the reality that the number of points reflected by different instances varies significantly, ranging from just a few to several hundreds or even thousands. However, due to the requirements of parallelization, all existing methods\cite{mppnet,msf,ptt} sample a fixed and large number of points, leading to considerable inefficiency. To obtain a better trade-off, we aim to represent the object using a limited number of notable points, referred to as focal tokens. As shown in \cref{fig1}(b), we oversample points just in the current frame, encode feature tokens and feed them into Single-frame Sequence Processing (SSP). In SSP, a simple but effective Adaptive Scaling mechanism is proposed for geometric feature interaction while dynamically acquiring focal tokens for sequence condensing. 

To handle the exponentially increasing computational load, we store only the selected notable points in memory bank, rather than the entire scene. Additionally, a reduced number of historical points are sampled to encode motion features, generating more compact and information-dense temporal sequences. 

In the Multi-frame Sequence Processing (MSP), we abandon the conventional result-level concatenation routine instead to propose an innovative grouped hierarchical fusion strategy. FASTer employs Adaptive Scaling to scale down each historical sequence while summarizing valid information. Additionally, an equal-step grouping strategy along with in Intra-Group Fusion (IGF) module are designed to refine the structure of temporal sequences. We do not explicitly separate spatial and temporal fusion instead to employ a united paradigm to progressively condense multi-frame sequences into a single long token sequence encompassing rich geometric and semantic information. Subsequently, a hierarchical decoder is introduced to summarize the outputs of SSP and MSP.

We conduct experiments on the Waymo Open Dataset\cite{waymo}, and the results demonstrate that our FASTer outperforms existing online detectors in terms of both effectiveness and efficiency, setting a new state-of-the-art (SOTA) in lidar-based 3D objection detection benchmark. Additionally, we conduct detailed ablation studies to validate the effectiveness of our meticulously designed components and find that FASTer offers several other notable advantages.

The main contributions of our work are as follows:
\begin{itemize}[label=\raisebox{-0.1\height}{\scalebox{1.0}{$\bullet$}},left=1em]
    \item We introduce the concept of the focal tokens and propose an Adaptive Scaling mechanism for focal token acquiring, combined with a lightweight point storage and encoding strategy, which enables the efficient processing of longer-sequence of point clouds.
    \item We propose a heuristic spatial-temporal fusion strategy, termed grouped hierarchical fusion, which asymptotically condenses the temporal sequences while seamlessly aggregating both spatial and temporal information, thereby comprehensively extracting the global context.
    \item We design comprehensive experiments to compare and analyze all region-guided methods. Our proposed FASTer achieves significant advantages in both performance and efficiency over existing temporal detectors, while maintaining enhanced flexibility and robustness.
\end{itemize}

\section{Related Work}

\begin{figure*}
    \centering
    \includegraphics[width=1.0\linewidth]{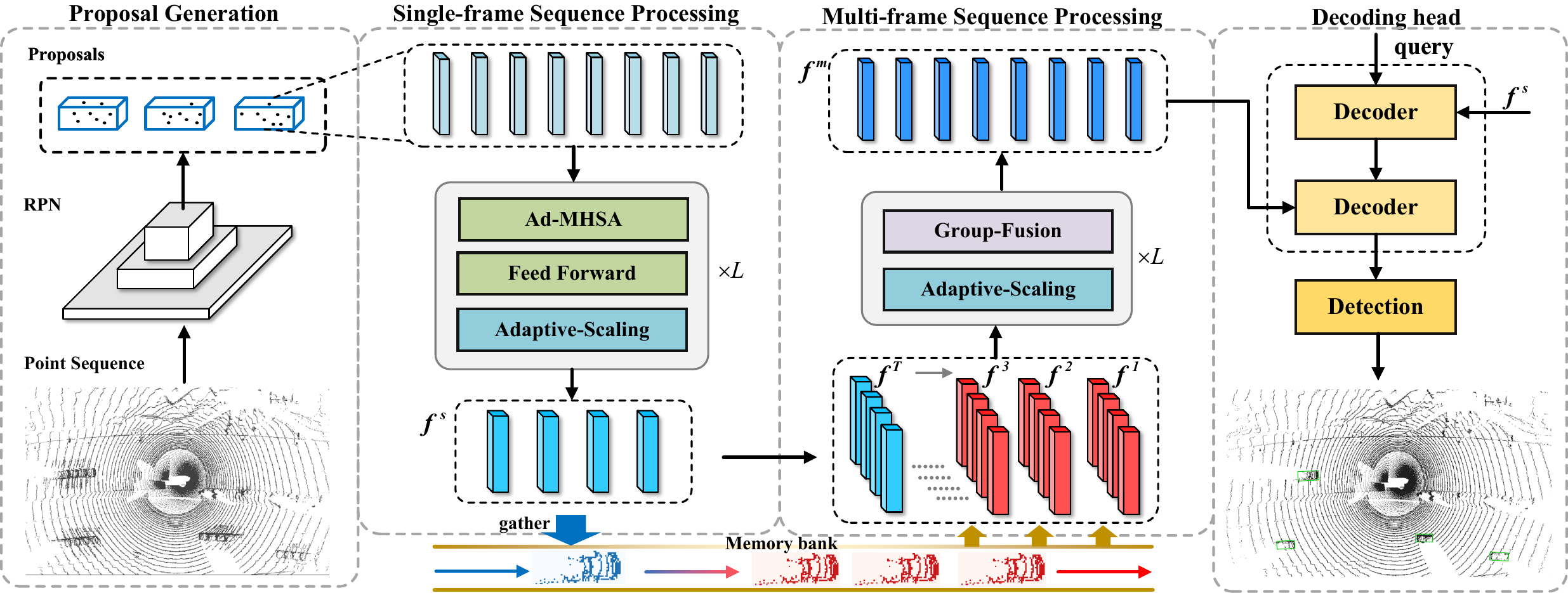}
    \caption{The overall architecture of \textbf{FASTer}. We begin by utilizing a region proposal network (RPN) to generate proposals. In the Single-frame Sequence Processing (SFP) illustrated in \cref{sec:CFE}, we randomly sample a sufficient number of points and apply adaptive scaling layers to extract geometric features while concurrently select notable points to store in the memory bank. In the Multi-frame Sequence Processing (MSP) illustrated in \cref{sec:MFF}, we just sample a reduced number of points and design a hierarchical fusion strategy to progressively condense the long temporal sequences into one single token sequence. Finally, a dual-layer decoder is introduced to aggregate the outputs of SSP and MSP. The blue and red cubes represent current and historical tokens respectively.}
    \label{fig:overview}
\end{figure*}
\subsection{3D Object Detection from Single Frame}

 Mainstream 3D object detectors follow a "feature extraction and detection" paradigm, corresponding to the two key components in the detection pipeline: the backbone and the detecting head. In recent years, numerous innovations have been proposed focusing on these two parts.

\paragraph{Feature Extraction.}
Point-based approaches\cite{pointrcnn,3dssd,votenet,IA-SSD,lidarrcnn} utilize set abstraction and local PointNet\cite{pointnet,pointnet++} to extract features directly from the point clouds, preserving fine-grained details. However, these methods are computationally intensive, limiting their applicability for processing large-scale ourdoor point clouds. By contrast, grid-based methods\cite{voxelnet,voxelrcnn,second,sst,voxelnext,voxeltransformer,dsvt,scatterformer,hednet,flatformer,pointpillars,swformer} partition the point cloud into discrete grids (voxels or pillars) and employ CNNs or transformer-like architectures for feature extraction. This grid-based approach achieves a favorable balance between effectiveness and efficiency, making it a mainstream choice.
\paragraph{Detection Head.}
Traditional detection heads (anchor-based\cite{second,pointpillars} or center-based\cite{centerpoint,centerformer}) are dense in design. Inspired by DETR in the 2D domain, some query-based heads\cite{seed,transfusion,cmt,focalformer3d} have been proposed to promote instance interaction. Additionally, two-stage, proposal-based heads are developed to further enhance performance. Furthermore, feature-level methods store the backbone's output (e.g., points\cite{pointrcnn}, grids\cite{voxelrcnn}, or BEV\cite{centerpoint}), and perform interpolation within proposed regions. CT3D\cite{ct3d}, however, discards feature maps instead to sample points from the ROI to capture the fine-grained contextual information. We refer to it as region-based method.

\subsection{3D Object Detection from Lidar Sequence}

Long-term point cloud sequences provide a more accurate and comprehensive view of 3D spatial states. Feature-based methods\cite{centerformer,mgtanet,fsd,3dman} employ shared feature extractors for each frame, followed by feature-level fusion. But aligning and interpolating features are sensitive and result in the loss of fine details. By contrast, region-based methods establish natural temporal connections between consecutive frames and better capture point cloud details. Offboard\cite{offboard} and MPPNet\cite{mppnet} link box proposals from adjacent frames to create trajectories, while MSF\cite{msf} proposes a motion-guided sequential fusion without prior boxes. PTT\cite{ptt} extends box trajectories longer by discarding historical points, but this relies on the high proposal recall rate and can loss essential semantic information from previous points, impacting the model performance.

Emphasizing the importance of both points and boxes, in this paper, we propose the highly lightweight FASTer, which utilizes a dynamic scaling mechanism to extend the region-based idea to longer sequences, achieving improvements in both effectiveness and efficiency.


\section{Methodology}

The overall architecture of our FASTer, as shown in \cref{fig:overview}, is composed of Proposal Generation, Single-frame Sequence Processing (SSP), Multi-frame Sequence Processing (MSP) and Decoding head. We initially utilize a Region Proposal Network (RPN) to generate proposals as previous works\cite{msf,ptt,mppnet}. 
In SSP, we provide a detailed description of our Adaptive Scaling mechanism, along with the strategies for focal points selecting and storage.
In \cref{sec:MFF}, we provide detailed representation of the proposed grouped hierarchical fusion strategy.
In \cref{sec:output}, we present the dual-decoding head and the loss functions of our FASTer.

\subsection{Single-frame Sequence Processing}
\label{sec:CFE}
\paragraph{Geometric Feature Encoding.}
To ensure adequate sampling in current frame, we randomly sample a sufficient number (denoted as $4K$) of points from the cylindrical region of each proposal, and any shortfall is padded with zero.
We then encode regional geometric features. Specifically, given sampled points $P^T=\{p_i^T\}_{i=1}^{4K}$ and box $B^T$ in current frame, the geometry embedding $ g^T $ is generated by calculating the offsets between each point $p^T_i$ and the 9 key points $\{b^T_j\}_{j=0}^{8}$ (8 corners and 1 center point) of the corresponding box, and then mapped to high-dimension space using a linear layer as \cite{ct3d}:
\begin{equation}
  \fontsize{10}{11.4}  g^T_i=\text{MLP}(\text{Concat}(\{p^T_i-b^T_j\}_{j=0}^8,r_i^T)),\; i=1,...,4K,
\label{geometry}
\end{equation}
where $ r_i^T$ represents extra point features. 

\paragraph{Adaptive Scaling.}
Previous studies\cite{IA-SSD,dbq} have revealed that not all points are equal, and foreground points require greater emphasis. We observe the reality that the number of reflected points varies widely across instances, from just a few to hundreds. The presence of background and padding points in the point sequence leads to unnecessary computational and storage overhead during scale-invariant inference. 
Inspired by token compression\cite{toc3d,evit,dynamicvit} in vision transformers, we interpret region-based methods as a variable-length sequence modeling problem and aim to dynamically compress the sequence length. 

\begin{figure}
    \centering
    \includegraphics[width=1.0\linewidth]{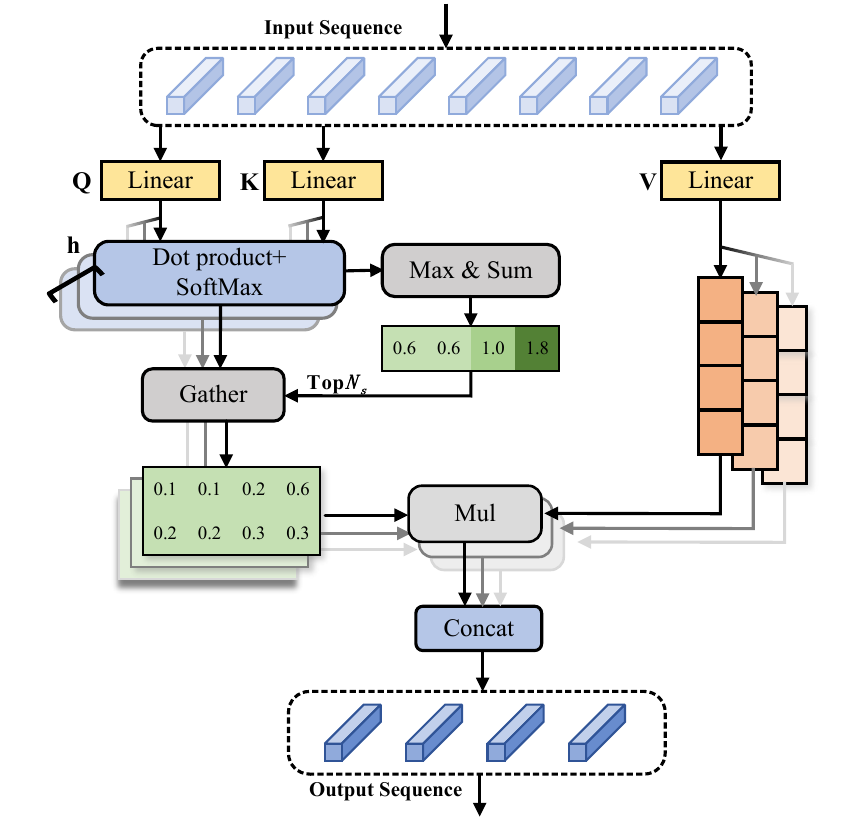}
    \caption{The architecture of Ad-MHSA, adaptively compresses long sequence while preserving the most valuable information.}
    \label{fig:ad-mhsa}
\end{figure}

However, unlike images where each pixel carries a clear semantic meaning, individual points cannot establish a direct relationship with the overall query and lacks explicit semantic supervision. As a result, class token attention\cite{evit} and score prediction\cite{dynamicvit} are not suitable. Therefore, we propose a simple yet effective Adaptive Multi-Head Self-Attention (Ad-MHSA) based on the attention map for sequence scaling.
As shown in \cref{fig:ad-mhsa}, given an input sequence $g^T\in \mathbb{R}^{4K\times D}$, where $d$ is the feature dimension, the attention map $A_h \in \mathbb{R}^{4K\times 4K}$of head $h$ is computed using dot product.


The contribution score $S\in [0,1]^{4K \times 1}$ to the global representation is computed as follows:
\begin{equation}
    S_i=\sigma(\sum_{j=1}^{4K} \underset{h\in [1,H]}{\text{max}} (\{{A_h^{ij}\}_{h=1}^{H}})),\; i=1,2,...,4K,
\end{equation}
where $\sigma$ denotes as the sigmoid function.

The indices $I\in\mathbb{R}^{N_s\times 1}$ of a fixed number of top $N_s$ (e.g.,$2K$) tokens with the highest scores are subsequently obtained.

Ultimately, the output $\hat{g}^{T} \in \mathbb{R}^{N_s \times d}$ of Ad-MHSA is focused on the selected tokens:
\begin{equation}
    \hat{g}^{T} = \text{Concat}(\{\text{gather}(A_h,I)\cdot V_h\}_{h=1}^H),
    \label{eq:update}
\end{equation}
where $V_h$ is the same as in \cite{attention}, and the gather function can be represented as follows:
\begin{equation}
     A^*=\text{gather}(A,I) \rightarrow  A^*[i,j] = A[I[i,j],j].
\end{equation}

Compared to existing token compression techniques\cite{evit,dynamicvit}, the rationale behind Ad-MHSA is that the scores and token selection are determined holistically, requiring no manual hyperparameters or additional supervision. This makes it more aligned with the characteristics of point cloud data. We refer to it as Adaptive Scaling.

The term \textbf{Adaptive} refers to that the focal tokens are learned by the network rather than manually defined. In the appendix, we provide a visualization of the token score distribution, which illustrates the transition from local to global understanding as the network recognizes instances. We apply Ad-MHSA progressively, reducing the token sequence length at each step, ultimately condensing the long sequence into a limited length (e.g., $K$). 



\paragraph{Focal Token Gathering and Storage.}
Some region-based methods\cite{msf,mppnet} face heavy storage demands for full point clouds across historical frames. In contrast, we store only the final focal points of each box in the memory pool. However, due to the overlap among proposals, one point may be sampled by different boxes. We construct unique indices to avoid duplication of individual points.

Specifically, during initial random sampling, assuming there are $M$ proposals and a total of $N$ unique points sampled, we denote this set as $P\in \mathbb{R}^{N\times 3}$, regardless of the sampling frequency. A matrix $I \in \mathbb{R}^{M \times 4K}$ is generated to record the indexes of points in $P$ for each proposal. And $I$ is updated during the token update in Eq.~\ref{eq:update}. From the index matrix $I^* \in \mathbb{R}^{M \times K}$ obtained from the final update, we extract the unique values, denoted as $\hat{I}\in \mathbb{R}^{\hat{N}\times 1}$ , where $\hat{N}$ represents the total number of focal points in the scene. The final focal points $\hat{P}\in \mathbb{R}^{\hat{N}\times 3}$ to be stored are obtained by:
\begin{equation}
    \hat{P} = \text{gather}(P,\hat{I}).
\end{equation}


We refer to the entire process above as adaptive sampling. In \cref{tab:point}, we report the quantities of complete point cloud, the proposal-region-level points after adaptive sampling. Obviously, through object-level sampling, a large number of background points are filtered out, while owing to adaptive sampling, the number of focal points to be stored is further compressed, rendering it nearly negligible compared with the complete scene.


\subsection{Multi-frame Sequence Processing}
\label{sec:MFF}
Building on existing region-based works\cite{msf,mppnet,ptt}, it is natural to extend FASTer for modeling longer temporal sequences. Utilizing the stored focal tokens, for preceding frames, we just need to sample and encode a number of points equivalent to the stored quantity for each box, thereby generating more compact and information-dense temporal sequences. However, fusing features from long sequences poses certain challenges. In the following paragraphs, we will give a detailed introduction to Multi-frame Sequence Processing (MSP), where we propose a progressive Grouped Hierarchical Fusion paradigm.

\paragraph{Motion Feature Encoding.}
We first generate the box trajectories as in\cite{ptt,mppnet}. To make the best use of focal points, if the maximum IOU between the current box and all boxes in the previous frame is below the given threshold, the proxy box projected with negative velocity serves as the substitute. We sample only $K$ points, which is equivalent to one-quarter of the sampling number in SSP. Thanks to the adaptive sampling and storage, there is no need for extensive coordinate transformations and point-box distance calculations, reducing the model's complexity from $O(FN)$ to nearly $O(N)$, where $F$ and $N$ represent the number of frames and full points respectively. 

We then encode the geometric features in each frame between the sampled points $g^t$ and the box $b^t$ as in Eq.~\ref{geometry}. Additionally, the motion features between historical points $p^t$ and the current frame box $b^T$ are obtained as follows:
\begin{equation}
    m^t_i=\text{MLP}(\text{Concat}(\{p^t_i-b^T_j\}_{j=0}^8,\Delta^t)),\; i=1,...,K,
\end{equation}
where $\Delta^t$ represents the time stamp at time $t$.

Then final feature of each token is obtained by summing the geometric feature and the motion feature:
\begin{equation}
    f^t_i = g^t_i + m^t_i,\; i=1,...,K.
\end{equation}

\begin{table}
  \centering
  \begin{tabular}{m{1.0cm}cm{1.2cm}m{1.2cm}m{1.2cm}}
    \toprule
 \centering    & Complete & Proposal region &  \centering Ours (4K=192) &    \quad Ours  (4K=128) \\
    \midrule
\centering    Num & 180k & 13.8k & \centering 2.6k & \quad 1.8k \\ 
    \bottomrule
  \end{tabular}
  \caption{The average number of points to be stored by different sampling methods.}
  \label{tab:point}
\end{table}
\setlength{\tabcolsep}{5pt}
\begin{table*}[ht]
\belowrulesep=0pt
 \aboverulesep=0pt
  \centering
  \small
  \begin{tabular}{c|c|cc|cc|cc|cc}
  \hline
    \toprule
\centering   
   \multirow{2}{*}{\centering Method} & \multirow{2}{*}{Frames} & \multicolumn{2}{c}{ALL (mAPH)} & \multicolumn{2}{|c}{VEH (AP/APH)}  & \multicolumn{2}{|c}{PED (AP/APH)} & \multicolumn{2}{|c}{CYC (AP/APH)}  \\
\cline{3-10} & & L1 & L2 &  L1 &  L2 & L1  & L2 & L1 & L2  \\
 
    \midrule
\centering PV-RCNN\cite{pvrcnn} &1 &69.63 & 63.33 & 77.51/76.89 & 68.98/68.41 & 75.01/65.65 & 66.04/57.61 & 67.81/66.35 & 65.39/63.98\\

\centering VoxSeT\cite{voxelset} & 1 & 72.24 & 66.22 & 74.50/74.03 & 65.99/65.56 & 80.03/72.42& 72.45/65.39& 71.56/70.29 & 68.95/67.73\\

\centering SST-1f\cite{sst} & 1 & -& -& 76.22/75.79 &68.04/67.64 &81.39/74.05& 72.82/65.93& -& - \\

\centering VoxelNext\cite{voxelnext} & 1 & 76.3 & 70.1 & 78.2/77.7 & 69.9/69.4 & 81.5/76.3 & 73.5/68.6 & 76.1/74.9 &73.3/72.2 \\

\centering FSD\cite{fsd} & 1 & 77.4 &70.8 & 79.2/78.8 &70.5/70.1 &82.6/77.3 & 73.9/69.1 & 77.1/76.0 & 74.4/73.3\\

\centering DSVT\cite{dsvt} & 1& 78.2 &72.1 & 79.7/79.3 & 71.4/71.0 & 83.7/78.9 & 76.1/71.5 & 77.5/76.5 & 74.6/73.7\\

\centering Clusterformer\cite{clusterformer} & 1 &79.0 & 72.3 & 79.8/79.3& 70.5/70.1 & 84.4/79.0 & 75.7/70.6 &80.0/78.7 & 77.4/76.2\\

\centering ScatterFormer\cite{scatterformer} & 1 & 79.3 & 73.6 & 80.4/79.9 & 72.8/72.4 & 84.1/81.1 &77.0/74.1 & 77.3/76.8 & 75.2/74.3\\

\midrule

\centering 3D-MAN & 16 & - & - &74.53/74.03 &67.61/67.14& 71.7/67.7& 62.6/59.0& - &-\\
 
\centering  SWFormer-3f\cite{swformer} &3 & -& - & 79.4/78.9 & 71.1/70.6 &82.9/79.0 & 74.8/71.1 & - & - \\
\centering CenterFormer\cite{centerformer}& 8 &   77.3 &73.7 &78.8/78.3 &74.3/73.8 & 82.1/79.3 & 77.8/75.0 & 75.2/74.4 & 73.2/72.3\\

\centering PillarNeXt-B\cite{pillarnext} & 3  & 80.04 & 74.45  & 80.58/80.08 &72.89/72.42 &85.04/82.11 &78.04/75.19 &78.92/77.93 &76.71/75.74 \\

\centering  MPPNet\cite{mppnet}& 4 &79.83& 74.22 &81.54/81.06 & 74.07/73.61 & 84.56/81.94 & 77.20/74.67 & 77.15/76.50 &75.01/74.38\\

\centering  MPPNet\cite{mppnet}& 16 &80.40 & 74.85 & 82.74/82.28 &75.41/74.96 & 84.69/82.25 & 77.43/75.06 & 77.28/76.66 & 75.13/74.52\\

\centering MSF\cite{msf} & 4 &80.20  &74.62 &81.36/80.87 & 73.81/73.35 &85.05/82.10 &77.92/75.11 &78.40/77.61  & 76.17/75.40 \\

\centering  MSF\cite{msf} &8 & 80.65 & 75.46 & 82.83/82.01  & 75.76/75.31 & 85.24/82.21  & 78.32/75.61 & 78.52/77.74 &  76.32/75.47\\

\centering  PTT\cite{ptt} &32 & 81.11 & 75.48 & 83.31/82.82  & 75.83/75.35 & 85.91/82.94  & 78.89/76.00 & 78.29/77.57 &  75.81/75.11\\

\centering PTT\cite{ptt} & 64 &81.32 & 75.71 & \textbf{83.71/83.21} &76.26/75.78 & 85.93/82.98 & 78.90/76.02 & 78.51/77.79 & 76.01/75.32\\

\midrule
\centering   FASTer(Ours)    &16      & 81.49 & 75.92 & 83.21/82.77  & 76.01/75.57  & 85.94/83.35  & 79.14/76.61 & 79.09/78.37 & 76.50/75.58\\

FASTer(Ours)    &32      & \textbf{81.62} & \textbf{76.06} & 83.53/83.02  & \textbf{76.32}/\textbf{75.81}  & \textbf{85.99}/\textbf{83.39}  & \textbf{79.17}/\textbf{76.62} & \textbf{79.18/78.45} & \textbf{76.66/75.75}\\

    \bottomrule
\hline
  \end{tabular}
  \caption{Performance comparison on the validation set of Waymo Open Dataset. The best result of each category are \textbf{bolded}.}
  \label{tab:result}
\end{table*}

\paragraph{Grouped Hierarchical Fusion.}
The reduction of the number of tokens per frame facilitates longer and more compact temporal sequences. However, feature extraction and fusion from long sequences are not trivial. Directly exchanging information among all tokens is neither feasible nor effective. Existing methods\cite{msf,mppnet} typically follow a routine of alternatively performing intra-frame and inter-frame fusions before concatenating the outputs decoded from each frame. We contend that the simple concatenation is not conducive to the interaction and aggregation of global information. Leveraging the Ad-MHSA, a more effective and rational grouped hierarchical fusion paradigm is proposed to enable the stable integration of global features. 

As shown in \cref{fig:overview}, given $T$ temporal sequences, each with a length of $K$, denoted as $F=\{f^t \in \mathbb{R}^{K \times D}\}_{t=1}^T$. We first employ the Ad-MHSA to extract geometric dependencies in each frame while scaling the sequence length. Subsequently, we divide the long temporal sequences into several groups along the temporal axis, denoted as $\hat{F}= \{\hat{F}^g \}_{g=1}^G$, where $G$ represents the number of groups. The indices of subsequences in each group are evenly spaced, (e.g., $\hat{F}^1=\{ f^1,f^5,f^9,f^{13} \} , \hat{F}^2 = \{ f^2,f^6,f^{10},f^{14} \}$ when $T=16$ and $G=4$ ), ensuring that each group maintains a roughly equal global significance and thus can be fed into the shared network.

Subsequently, we need to aggregate the dependencies within each group. A straightforward approach is to conbine all sequences in each group, but this results in the loss of temporal information and renders our work a coarse stacking. Therefore, we design a more reasonable Intra-Group Fusion (IGF) module. 
\begin{figure}
    \centering
    \includegraphics[width=1.0\linewidth]{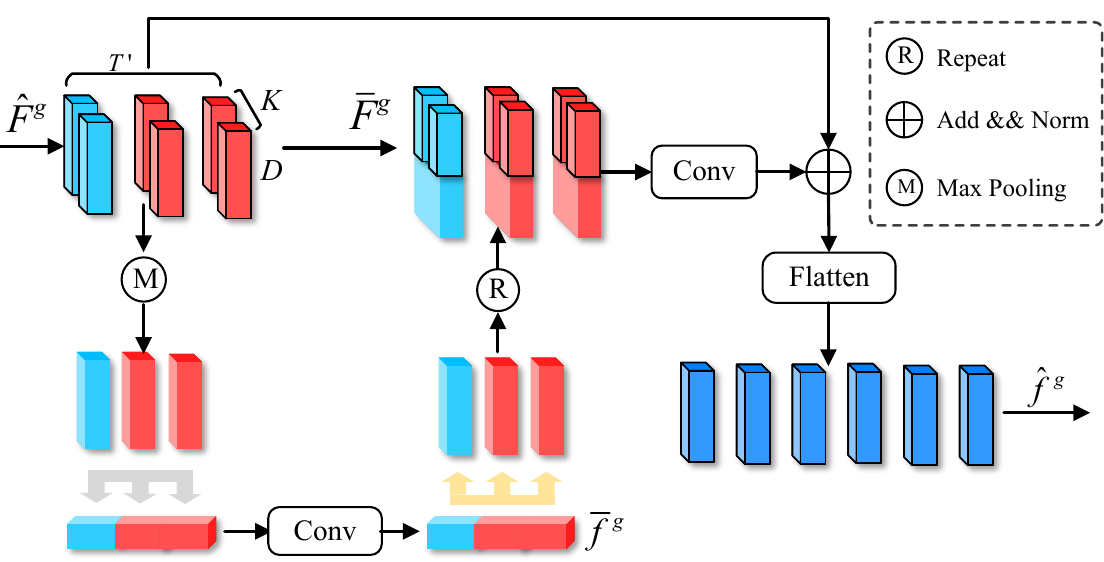}
    \caption{The architecture of Intra-Group Fusion, taking $T'$ sequences as input and combining them to output a single sequence.}
    \label{fig:group-fusion}
\end{figure}

As shown in \cref{fig:group-fusion}, with the input subsequences $\hat{F}^g\in \mathbb{R}^{T' \times K \times D}$ of group $g$, where $T'=T/G $, IGF performs max pooling within each sequence, followed by concatenation along the feature channels to obtain an integral representation:
\begin{equation}
    \bar{f}^g = \text{Conv}(\text{Concat}(\text{MaxPool}(\hat{F}^g))).
\end{equation}

$\bar{f}^g$ is forwardly partitioned into tokens of shape $T'\times D$, which are replicated $K$ times and concatenated to $\hat{F}^g$ along the feature channels, the result, denoted by $\bar{F}^g$, can be obtained by
\begin{equation}
    \bar{F}^g = \text{Concat}(\hat{F}^g,\text{Repeat}(\text{Split}(\bar{f}^g))).
\end{equation}

Subsequently, a channel compression convolution with shared weights and a residual connection layer are utilized, after which the output is flattened to one sequence, denoted as $\hat{f}^g \in\mathbb{R}^{(T'\times K)\times D}$:
\begin{equation}
\hat{f}^g = \text{Flatten}(\text{Norm}(\hat{F}^g+\text{Conv}(\bar{F}^g))).
\end{equation}

IGF processes all groups in parallel, reconstructing the $T$ sequences into $G$ groups to achieve temporal compression. Ad-MHSA and IGF are executed alternately, maintaining a relatively stable token count per sequence to balance computation overhead, ultimately condensing the long temporal sequences into a single one enriched with dense geometric and temporal information.

\subsection{Decoder and Loss}
\label{sec:output}
To fully exploit the capabilities of SSP and MSP, we propose a dual-layer decoder head where a learnable embedding $q\in \mathbb{R}^{1\times D}$ is utilized as query and passed through two sequential layers of standard transformer decoders. The output $f^c$ from SSP and $f^m$ from MSP serve as the key and value for the first and second layers of decoder, respectively. The complete decoding head is formalized as follows:
\begin{equation}
\begin{split}
    q^s&=\text{FFN}(q+\text{Attention}(q,f^s,f^s)),\\
    q^m&=\text{FFN}(q^s+\text{Attention}(q^s,f^m,f^m)).
\end{split}
\end{equation}

The final output $q^m$ is sent to the detection head for bounding box refinement.


The overall loss $\mathcal{L}_{\text{all}} $ is composed of both the confidence loss $ \mathcal{L}_{\text{conf}}$ and the regression loss $\mathcal{L}_{\text{reg}}$ as follows:
\begin{equation}
    \mathcal{L}_{\text{all}} = \mathcal{L}_{\text{conf}}+\alpha\mathcal{L}_{\text{reg}}.
\end{equation}
where $\alpha$ is the hyper-parameter for balancing different losses. We employ cross binary loss as $ \mathcal{L}_{conf}$ and the same regression loss as other region-based methods\cite{ct3d,mppnet,msf,ptt}. In the training stage, the output of each decoding layer is supervised immediately to help the model learn. While during the test phase, only the output decoded from MSP is utilized as the final detection result.


    
    
    

\section{Experiment}
\subsection{Dataset and Implementation}
\paragraph{Dataset.}
We evaluate our FASTer on Waymo Open Dataset (WOD)\cite{waymo}, one of the most popular dataset for 3D Object Detection. WOD consists of 1,150 sequences, which are further divided into 798 for training, 202 for validating, and 150 for testing. Each sequence is obtained by five 64-line lidars scanning for 20 seconds at 10Hz. The official evaluation metrics are 3D mean average precision (mAP) and 3D mAP weighted by heading accuracy (mAPH). There are three categories to be detected: Vehicle, Pedestrian and Cyclist. Each category is categorized into LEVEL1 and LEVEL2 according to the number of points of objects, containing at least 5 points and 1 point respectively.

\paragraph{Implementation Details.}
Following established works\cite{mppnet,msf,ptt}, we adopt 4-frame CenterPoint\cite{centerpoint} as the Region Proposal Network (RPN). Our FASTer is trained for 6 epochs with a batch size of 5, utilizing the Adam optimizer and an initial learning rate of $3e-4$. The One-Cycle learning rate scheduler is utilized. The IOU match threshold is set to 0.5 for proposal trajectory generation. Box augmentation outlined in \cite{pointrcnn} is also applied. The point sampling constant $K$ is set to 48, i.e. we sample $4K=192$ points in SSP and $K=48$ points of each preceding frame in MSP. For the Ad-MHSA, we utilize 8 attention heads with a feature dimension of 256.

Notably, the focal points are acquired by learning and cannot be obtained in advance. Therefore, we design a staged training strategy to facilitate learning. Additionally, to reduce the reliance on the Region Proposal Network (RPN), we develop an Extra Point Augmentation (EPA) to enhance training. Experimental results in \cref{tab:ablation} demonstrate the generalization improvement brought by EPA. More details can be found in the appendix.

\setlength{\tabcolsep}{5.0pt}
\begin{table}[]
\belowrulesep=0pt
 \aboverulesep=0pt
    \centering
    \begin{tabular}{c |c |c  c }
    \toprule
\rule{0pt}{2.0ex}\vspace*{0.00ex}     \multirow{2}{*}{\centering \makecell{Method}} & \multirow{2}{*}{\centering \makecell{Frame}}  & \multicolumn{2}{c}{ mAP   /   mAPH}     \\
         & &    L1  & L2\\
    \midrule
    SST\cite{sst} &3 & 80.0/78.3 & 74.4/72.8\\
    MPPNet\cite{mppnet}&  16 & 81.83/70.59  &  77.60/75.67          \\ 
    ScatterFormer\cite{scatterformer} & 4 & 82.11/80.71 &  77.25/75.91\\

     CenterFormer\cite{centerformer} & 16 & 82.26/80.91   & 77.60/76.29  \\

     MSF\cite{msf} &  8 &  83.11/81.74 &   78.30/76.96         \\ 
     \midrule
    FASTer(Ours) & 16  &   83.28/81.95 & 78.53/77.21  \\
     FASTer(Ours) & 32  &   \textbf{83.55/82.23} & \textbf{78.82/77.54}  \\

    \bottomrule
    \end{tabular}
    \caption{Performance comparison on the test set of Waymo Open Dataset. The best result of each category are \textbf{bolded}.}
    \label{tab:test}
\end{table}

\begin{table}[]
\belowrulesep=0pt
 \aboverulesep=0pt
    \centering
    \begin{tabular}{c | c | c c c}
    \toprule
    
    \rule{0pt}{2.5ex}\vspace*{0.2ex} Model   &Frame&  mAPH(L2)$\mathbf{\uparrow}$   &   Delay$\mathbf{\downarrow}$  & Memory$\mathbf{\downarrow}$\\
    \midrule
     MPPNet\cite{mppnet} & 4 & 74.22 &  332ms    &  4153M   \\
     MSF\cite{msf}  &  4 & 74.62 &   101ms       &   3752M     \\ 
     MSF\cite{msf} &  8 &  75.46 &  400ms &    6083M         \\ 
     PTT\cite{ptt} &  32 & 75.48 &  99 ms        & 6938M     \\
     \midrule
     Ours & 16  &   \textbf{75.92} & \textbf{ 75 ms } & \textbf{2856M} \\
    \bottomrule
    \end{tabular}
    \caption{Comparison of effectiveness and efficiency of different models. }
    \label{tab:efficiency}
\end{table}

\subsection{Results on WOD}
\paragraph{Effectiveness Comparison}
In \cref{tab:result}, we report the performance comparison between our FASTer and various lidar-based state-of-the-art (SOTA) methods. Generally, multi-frame approaches outperform single-frame ones. Our 16-frame FASTer surpasses the strongest one-stage detector ScatterFormer\cite{scatterformer} by 2.19\% mAPH and 2.32\% mAPH on LEVEL1 and LEVEL2. Compared with multi-frame methods, FASTer significantly outperforms all region-based approaches, especially in detecting small objects. With 16 frames as input and a similar grouping design, out FASTer outperforms MPPNet\cite{mppnet} by 1.09\% mAPH on LEVEL1 and 1.07\% mAPH on LEVEL2. In \cref{tab:test}, we present the results of different models on the WOD test set. Our FASTer consistently and significantly outperforms other arts, demonstrating its strong generalizability.

 To fully unlock the potential of FASTer, we replace the CNN backbone with more powerful transformer-based backbones\cite{dsvt,scatterformer}, and the results can be found in the appendix. 

\paragraph{Efficiency Comparison.}
In \cref{tab:efficiency}, we compare the efficiency of various region-based methods using the same proposals. As the name suggests, our FASTer demonstrates high efficiency and resource-friendliness, as the lightweight model surpasses the performance of MSF (8-frame)\cite{msf}, while maintaining lower latency and memory usage than MSF (4-frame). We attribute this to our adaptive scaling, focal token gathering, storage and hierarchical fusion strategies, which significantly enhances resource utilization. The results are tested on an NVIDIA 3090 GPU.
\noindent

\subsection{Ablation Studies}
In this Section, we design detailed ablation experiments to validate the contribution of each component and the robustness of our FASTer. By default, we report the mAPH of our 16-frame model on the WOD validation set.

\begin{table}[]
\belowrulesep=0pt
 \aboverulesep=0pt
    \centering
    \begin{tabular}{c c c c | c c}
    \toprule
\rule{0pt}{2.5ex}\vspace*{0.2ex}     SD   & MD     & IGF & ME      &    mAPH(L1)     & mAPH(L2)    \\
    \midrule
    \ding{51} & \ding{51}  & \ding{51} & \ding{51} & 81.49     &  75.92     \\
    \ding{55} & \ding{51}  & \ding{51} & \ding{51} & 81.01(\textcolor{blue}{-0.48})    &  75.51(\textcolor{blue}{-0.41})     \\ 
    \ding{51} & \ding{55}  & \ding{51} & \ding{51} & 79.02(\textcolor{blue}{-2.47})  &  73.19(\textcolor{blue}{-2.73})      \\ 
    \ding{51} & \ding{51}  & \ding{55} & \ding{51} &  81.06(\textcolor{blue}{-0.43})     &   75.45(\textcolor{blue}{-0.47})     \\ 
     \ding{51} & \ding{51}  & \ding{51} & \ding{55} &  80.31(\textcolor{blue}{-1.18})     &   74.72(\textcolor{blue}{-1.20})     \\ 

    \bottomrule
    \end{tabular}
    \caption{Ablation experiments for the core components. \textbf{SD} and \textbf{MD} mean the decoder layers of SSP and MSP respectively while \textbf{ME} means the motion encoding.}
    \label{tab:ablation}
\end{table}

\begin{table}[]
\belowrulesep=0pt
 \aboverulesep=-1.5pt
    \centering
    \begin{tabularx}{0.48\textwidth}{c >{\centering\arraybackslash}X | >{\centering\arraybackslash}X >{\centering\arraybackslash}X  >{\centering\arraybackslash}X} 
    \toprule
\rule{0pt}{2.5ex}\vspace*{0.2ex} Config & Concat  & Vehicle & Pedes & Cyclist      \\
    \midrule
   S-T Fusion   & \ding{55}&   75.57&  76.61      & 75.58    \\
   Cross Atten\cite{mppnet} &\ding{51} & 75.22  & 76.03 & 75.35  \\ 
    BiFA\cite{msf}  & \ding{51}& 75.01 &    75.97 &  75.25     \\ 
     
    \bottomrule
    \end{tabularx}
    \caption{Comparative experiments of different temporal-spatial fusion paradigms. \textbf{Concat} means that all sequences are decoded separately and then concatenated. APH on Level2 is reported.}
    \label{tab:st-fusion}
\end{table}

\paragraph{Effectiveness of Core Components.} 
As shown in rows 1 and 2 of \cref{tab:ablation}, without the SSP and MSP decoding layers, mAPH on Level2 drops by 0.41\% and 2.73\%, respectively. This demonstrates that temporal information is crucial and that the current frame and the history frame should be decoupled.

As shown in the row 3 of \cref{tab:ablation}, without Intra-Group-Fusion module as a bridge for compressing long sequences, FASTer suffers from the performance degradation of 0.43\% on Level1 and 0.47\% on Level2. This shows that simple combination do indeed lose a significant amount of valuable information and demonstrates the effectiveness of IGF. We also observed that the motion feature encoding brings 1.2\% performance improvement.


\paragraph{Effectiveness of S-T Fusion.}
To validate the effectiveness of our S-T Fusion paradigm, we design ablation experiments where only the first IGF is retained in MSP, followed by multiple layers of either BiFA in \cite{msf} or cross-attention inspired by \cite{mppnet}. The decoding and concatenation mechanisms remained unchanged in these setups. As shown in \cref{tab:st-fusion}, the replacement of S-T Fusion results in notable performance degradation, underscoring the importance of global modeling and hierarchical fusion.
\begin{table}[]
\belowrulesep=0pt
 \aboverulesep=0pt
    \centering
    \begin{tabular}{c c c  c c}
    \toprule
\rule{0pt}{2.5ex}\vspace*{0.2ex}     \multirow{2}{*}{\centering \makecell{Strategies of\\Scaling}}   &Vehicle & Pedestrian  &  Cyclist   \\
   \cline{2-4}  \rule{0pt}{2.5ex}\vspace*{0.2ex} &         L1 / L2          &   L1 / L2  & L1 / L2\\
    \midrule
    Adaptive  &   82.77/75.57 &83.35/76.61& 78.37/75.58       \\
    Supervised &   82.18/74.85   & 83.22/76.50  & 77.96/75.27 \\
    Atten Mask &  81.95/74.60 & 83.21/76.50  & 77.80/75.12 \\
    Random   &    81.45/74.11& 82.79/76.11 & 77.63/74.97     \\

    \bottomrule
    \end{tabular}
    \caption{Comparison of the effect of different scaling strategies.}
    \label{tab:admhsa}
\end{table}
\begin{table}[]
\centering
\begin{minipage}{0.5\textwidth}  
\centering
\belowrulesep=0pt
\aboverulesep=0pt
\begin{tabular}{>{\centering\arraybackslash}p{1.8cm}>{\centering\arraybackslash}p{2.5cm}>{\centering\arraybackslash}p{2.5cm}}
\toprule
\rule{0pt}{2.5ex} \vspace*{0.2ex}  \multirow{2}{*}{\centering \makecell{Inputted\\Frames}} & L1 & L2  \\
\cline{2-3} \rule{0pt}{2.5ex} \vspace*{0.2ex} & mAP / mAPH & mAP / mAPH \\
\hline
4  & 81.94 / 80.58 & 76.30 / 74.98 \\
8  & 82.34 / 81.03 & 76.78 / 75.51 \\
16 & 82.74 / 81.49 & 77.22 / 75.92 \\
24 & 82.83 / 81.56 & 77.31 / 76.00 \\
32 & 82.90 / 81.62 & 77.38 / 76.06 \\
\bottomrule
\end{tabular}
\caption{Comparison of different numbers of historical frames.}
\label{tab:frame}
\end{minipage}
\end{table}
\paragraph{Effectiveness of Adaptive Scaling.}
In \cref{tab:admhsa}, we compare various token compression methods. The results reveal that all conscious focal token selecting outperform random sampling, while our adaptive scaling achieves superior performance compared to in-box supervision and attention masking, which we attribute to their focus on local tokens, neglecting the global representation. Further details can be found in the appendix.

\paragraph{Analysis about the length of temporal sequences.}
By rational grouping, in \cref{tab:frame}, we compare the impact of varying frame counts on FASTer. The results show that as the input frame count increases from 4 to 32, FASTer's performance steadily improves, demonstrating its robustness. Moreover, with the same number of frames, FASTer consistently outperforms other competitors\cite{msf,mppnet,ptt}.

\begin{figure}
    \centering
    \includegraphics[width=1.0\linewidth]{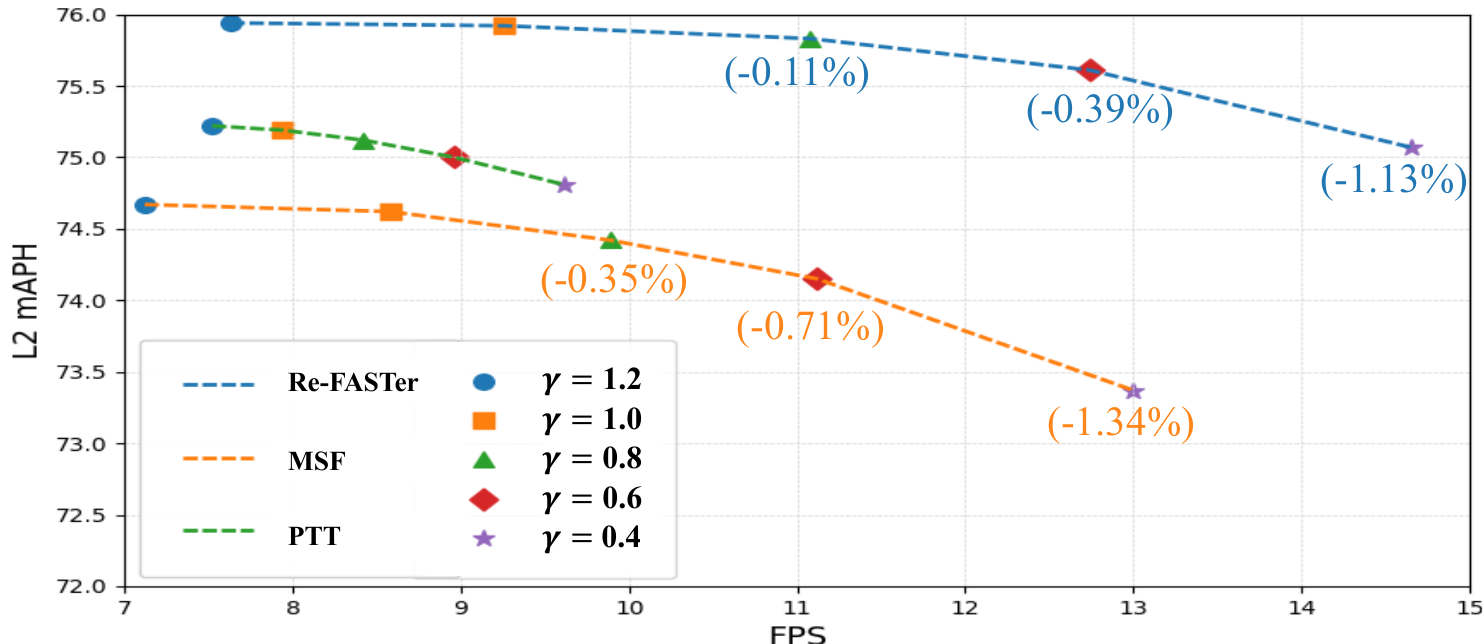}
    \caption{Sensitivity Comparison of several region-based methods with respect to the points sampling ratio $\gamma$.}
    \label{fig:fps}
\end{figure}

\paragraph{Sensitivity analysis of the number of sampled points.} 
We set the default number of sampled points as the baseline for each model and gradually adjust the sampling count by a factor of $\gamma$, with the sensitivity analysis presented in \cref{fig:fps}. We observe that as $\gamma$ drops, both FASTer and MSF\cite{msf} obtain significant acceleration in inference, but FASTer exhibits a milder drop in performance, indicating that instances can be effectively represented with fewer focal tokens. This property allows FASTer to maintain competitive performance even under hardware constraints.

\begin{figure}
    \centering
    \includegraphics[width=1.0\linewidth]{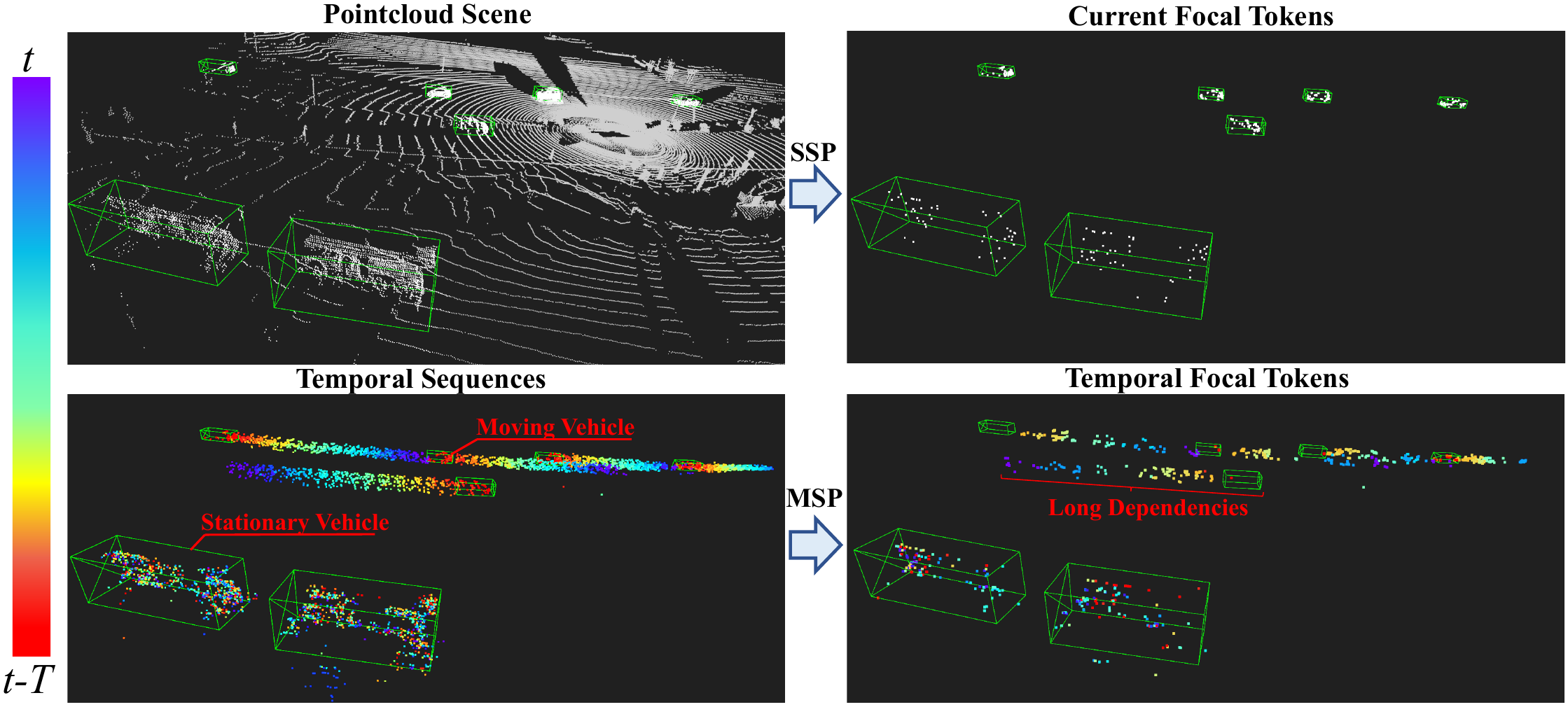}
    \caption{Visualization of focal tokens acquired by SSP and MSP. We differentiate the temporal sequences using different colors, while the ground truth boxes are represented in \textcolor{green}{green}.}
    \label{fig:vis_scene}
\end{figure}
\subsection{Visualization of Focal Tokens.}

To demonstrate the effectiveness of the focal tokens, we visualize the focal tokens obtained by SSP and MSP in a representative scene in \cref{fig:vis_scene}. It can be observed that the focal tokens are predominantly located on the surface of instances and exhibit a scattered distribution along both the spatial and temporal dimensions, which confirms that FASTer successfully captures global dependencies. Furthermore, an interesting phenomenon is noted: in the temporal sequence of \textbf{moving instances}, the historical tokens contribute a \textbf{larger proportion}, suggesting that they should be assigned more consideration.


\section{Conclusion}
In this paper, we extend the concept of region-based fusion by conceptualizing it as a variable-length sequence modeling problem. We propose FASTer, a lightweight, lidar-based multi-frame detector. We demonstrate that objects can be effectively represented with a limited number of focal tokens and propose the principle of adaptive scaling to acquire focal tokens, dramatically alleviating the exponential computational and storage complexity issues. To overcome the limitations of the \textbf{fusion-concatenate} approach seen in other methods, FASTer introduces an effective grouped hierarchical fusion paradigm and a dual-decoding mechanism to more coherently aggregate spatiotemporal information, facilitating the global-level fusion. FASTer achieves outstanding performance and efficiency on Waymo Open Dataset against other methods. We further design comprehensive experiments to compare all region-based temporal detectors, demonstrating that FASTer exhibits superior flexibility and robustness.


{
    \small
    \bibliographystyle{ieeenat_fullname}
    \bibliography{main}

\begin{thebibliography}{45}
\providecommand{\natexlab}[1]{#1}
\providecommand{\url}[1]{\texttt{#1}}
\expandafter\ifx\csname urlstyle\endcsname\relax
  \providecommand{\doi}[1]{doi: #1}\else
  \providecommand{\doi}{doi: \begingroup \urlstyle{rm}\Url}\fi

\bibitem[Bai et~al.(2022)Bai, Hu, Zhu, Huang, Chen, Fu, and Tai]{transfusion}
Xuyang Bai, Zeyu Hu, Xinge Zhu, Qingqiu Huang, Yilun Chen, Hongbo Fu, and Chiew-Lan Tai.
\newblock Transfusion: Robust lidar-camera fusion for 3d object detection with transformers.
\newblock In \emph{Proceedings of the IEEE/CVF conference on computer vision and pattern recognition}, pages 1090--1099, 2022.

\bibitem[Chen et~al.(2022)Chen, Shi, Zhu, Cheung, Xu, and Li]{mppnet}
Xuesong Chen, Shaoshuai Shi, Benjin Zhu, Ka~Chun Cheung, Hang Xu, and Hongsheng Li.
\newblock Mppnet: Multi-frame feature intertwining with proxy points for 3d temporal object detection.
\newblock In \emph{European Conference on Computer Vision}, pages 680--697. Springer, 2022.

\bibitem[Chen et~al.(2023{\natexlab{a}})Chen, Liu, Zhang, Qi, and Jia]{voxelnext}
Yukang Chen, Jianhui Liu, Xiangyu Zhang, Xiaojuan Qi, and Jiaya Jia.
\newblock Voxelnext: Fully sparse voxelnet for 3d object detection and tracking.
\newblock In \emph{Proceedings of the IEEE/CVF Conference on Computer Vision and Pattern Recognition}, pages 21674--21683, 2023{\natexlab{a}}.

\bibitem[Chen et~al.(2023{\natexlab{b}})Chen, Yu, Chen, Lan, Anandkumar, Jia, and Alvarez]{focalformer3d}
Yilun Chen, Zhiding Yu, Yukang Chen, Shiyi Lan, Anima Anandkumar, Jiaya Jia, and Jose~M Alvarez.
\newblock Focalformer3d: focusing on hard instance for 3d object detection.
\newblock In \emph{Proceedings of the IEEE/CVF International Conference on Computer Vision}, pages 8394--8405, 2023{\natexlab{b}}.

\bibitem[Deng et~al.(2021)Deng, Shi, Li, Zhou, Zhang, and Li]{voxelrcnn}
Jiajun Deng, Shaoshuai Shi, Peiwei Li, Wengang Zhou, Yanyong Zhang, and Houqiang Li.
\newblock Voxel r-cnn: Towards high performance voxel-based 3d object detection.
\newblock In \emph{Proceedings of the AAAI conference on artificial intelligence}, pages 1201--1209, 2021.

\bibitem[Fan et~al.(2022)Fan, Pang, Zhang, Wang, Zhao, Wang, Wang, and Zhang]{sst}
Lue Fan, Ziqi Pang, Tianyuan Zhang, Yu-Xiong Wang, Hang Zhao, Feng Wang, Naiyan Wang, and Zhaoxiang Zhang.
\newblock Embracing single stride 3d object detector with sparse transformer.
\newblock In \emph{Proceedings of the IEEE/CVF conference on computer vision and pattern recognition}, pages 8458--8468, 2022.

\bibitem[He et~al.(2022)He, Li, Li, and Zhang]{voxelset}
Chenhang He, Ruihuang Li, Shuai Li, and Lei Zhang.
\newblock Voxel set transformer: A set-to-set approach to 3d object detection from point clouds.
\newblock In \emph{Proceedings of the IEEE/CVF Conference on Computer Vision and Pattern Recognition (CVPR)}, pages 8417--8427, 2022.

\bibitem[He et~al.(2023)He, Li, Zhang, Li, and Zhang]{msf}
Chenhang He, Ruihuang Li, Yabin Zhang, Shuai Li, and Lei Zhang.
\newblock Msf: Motion-guided sequential fusion for efficient 3d object detection from point cloud sequences.
\newblock In \emph{Proceedings of the IEEE/CVF Conference on Computer Vision and Pattern Recognition}, pages 5196--5205, 2023.

\bibitem[He et~al.(2024)He, Li, Zhang, and Zhang]{scatterformer}
Chenhang He, Ruihuang Li, Guowen Zhang, and Lei Zhang.
\newblock Scatterformer: Efficient voxel transformer with scattered linear attention.
\newblock \emph{arXiv preprint arXiv:2401.00912}, 2024.

\bibitem[Huang et~al.(2024)Huang, Lyu, Yang, and Tsai]{ptt}
Kuan-Chih Huang, Weijie Lyu, Ming-Hsuan Yang, and Yi-Hsuan Tsai.
\newblock Ptt: Point-trajectory transformer for efficient temporal 3d object detection.
\newblock In \emph{Proceedings of the IEEE/CVF Conference on Computer Vision and Pattern Recognition}, pages 14938--14947, 2024.

\bibitem[Koh et~al.(2023)Koh, Lee, Lee, Kim, and Choi]{mgtanet}
Junho Koh, Junhyung Lee, Youngwoo Lee, Jaekyum Kim, and Jun~Won Choi.
\newblock Mgtanet: Encoding sequential lidar points using long short-term motion-guided temporal attention for 3d object detection.
\newblock In \emph{Proceedings of the AAAI Conference on Artificial Intelligence}, pages 1179--1187, 2023.

\bibitem[Lang et~al.(2019)Lang, Vora, Caesar, Zhou, Yang, and Beijbom]{pointpillars}
Alex~H Lang, Sourabh Vora, Holger Caesar, Lubing Zhou, Jiong Yang, and Oscar Beijbom.
\newblock Pointpillars: Fast encoders for object detection from point clouds.
\newblock In \emph{Proceedings of the IEEE/CVF conference on computer vision and pattern recognition}, pages 12697--12705, 2019.

\bibitem[Li et~al.(2023)Li, Luo, and Yang]{pillarnext}
Jinyu Li, Chenxu Luo, and Xiaodong Yang.
\newblock Pillarnext: Rethinking network designs for 3d object detection in lidar point clouds.
\newblock In \emph{Proceedings of the IEEE/CVF Conference on Computer Vision and Pattern Recognition}, pages 17567--17576, 2023.

\bibitem[Li et~al.(2021)Li, Wang, and Wang]{lidarrcnn}
Zhichao Li, Feng Wang, and Naiyan Wang.
\newblock Lidar r-cnn: An efficient and universal 3d object detector.
\newblock In \emph{Proceedings of the IEEE/CVF conference on computer vision and pattern recognition}, pages 7546--7555, 2021.

\bibitem[Liang et~al.(2022)Liang, Ge, Tong, Song, Wang, and Xie]{evit}
Youwei Liang, Chongjian Ge, Zhan Tong, Yibing Song, Jue Wang, and Pengtao Xie.
\newblock Not all patches are what you need: Expediting vision transformers via token reorganizations.
\newblock \emph{arXiv preprint arXiv:2202.07800}, 2022.

\bibitem[Liu et~al.(2023)Liu, Yang, Tang, Yang, and Han]{flatformer}
Zhijian Liu, Xinyu Yang, Haotian Tang, Shang Yang, and Song Han.
\newblock Flatformer: Flattened window attention for efficient point cloud transformer.
\newblock In \emph{Proceedings of the IEEE/CVF Conference on Computer Vision and Pattern Recognition}, pages 1200--1211, 2023.

\bibitem[Liu et~al.(2024)Liu, Hou, Ye, Wang, Wang, and Bai]{seed}
Zhe Liu, Jinghua Hou, Xiaoqing Ye, Tong Wang, Jingdong Wang, and Xiang Bai.
\newblock Seed: A simple and effective 3d detr in point clouds.
\newblock \emph{arXiv preprint arXiv:2407.10749}, 2024.

\bibitem[Luo et~al.(2018{\natexlab{a}})Luo, Yang, and Urtasun]{faf}
Wenjie Luo, Bin Yang, and Raquel Urtasun.
\newblock Fast and furious: Real time end-to-end 3d detection, tracking and motion forecasting with a single convolutional net.
\newblock In \emph{Proceedings of the IEEE conference on Computer Vision and Pattern Recognition}, pages 3569--3577, 2018{\natexlab{a}}.

\bibitem[Luo et~al.(2018{\natexlab{b}})Luo, Yang, and Urtasun]{fsd}
Wenjie Luo, Bin Yang, and Raquel Urtasun.
\newblock Fast and furious: Real time end-to-end 3d detection, tracking and motion forecasting with a single convolutional net.
\newblock In \emph{Proceedings of the IEEE conference on Computer Vision and Pattern Recognition}, pages 3569--3577, 2018{\natexlab{b}}.

\bibitem[Mao et~al.(2021)Mao, Xue, Niu, Bai, Feng, Liang, Xu, and Xu]{voxeltransformer}
Jiageng Mao, Yujing Xue, Minzhe Niu, Haoyue Bai, Jiashi Feng, Xiaodan Liang, Hang Xu, and Chunjing Xu.
\newblock Voxel transformer for 3d object detection.
\newblock In \emph{Proceedings of the IEEE/CVF international conference on computer vision}, pages 3164--3173, 2021.

\bibitem[Pei et~al.(2023)Pei, Zhao, Li, Ma, Zhang, and Pu]{clusterformer}
Yu Pei, Xian Zhao, Hao Li, Jingyuan Ma, Jingwei Zhang, and Shiliang Pu.
\newblock Clusterformer: Cluster-based transformer for 3d object detection in point clouds.
\newblock In \emph{Proceedings of the IEEE/CVF International Conference on Computer Vision (ICCV)}, pages 6664--6673, 2023.

\bibitem[Qi et~al.(2017{\natexlab{a}})Qi, Su, Mo, and Guibas]{pointnet}
Charles~R Qi, Hao Su, Kaichun Mo, and Leonidas~J Guibas.
\newblock Pointnet: Deep learning on point sets for 3d classification and segmentation.
\newblock In \emph{Proceedings of the IEEE conference on computer vision and pattern recognition}, pages 652--660, 2017{\natexlab{a}}.

\bibitem[Qi et~al.(2017{\natexlab{b}})Qi, Yi, Su, and Guibas]{pointnet++}
Charles~Ruizhongtai Qi, Li Yi, Hao Su, and Leonidas~J Guibas.
\newblock Pointnet++: Deep hierarchical feature learning on point sets in a metric space.
\newblock \emph{Advances in neural information processing systems}, 30, 2017{\natexlab{b}}.

\bibitem[Qi et~al.(2019)Qi, Litany, He, and Guibas]{votenet}
Charles~R Qi, Or Litany, Kaiming He, and Leonidas~J Guibas.
\newblock Deep hough voting for 3d object detection in point clouds.
\newblock In \emph{proceedings of the IEEE/CVF International Conference on Computer Vision}, pages 9277--9286, 2019.

\bibitem[Qi et~al.(2021)Qi, Zhou, Najibi, Sun, Vo, Deng, and Anguelov]{offboard}
Charles~R Qi, Yin Zhou, Mahyar Najibi, Pei Sun, Khoa Vo, Boyang Deng, and Dragomir Anguelov.
\newblock Offboard 3d object detection from point cloud sequences.
\newblock In \emph{Proceedings of the IEEE/CVF Conference on Computer Vision and Pattern Recognition}, pages 6134--6144, 2021.

\bibitem[Rao et~al.(2021)Rao, Zhao, Liu, Lu, Zhou, and Hsieh]{dynamicvit}
Yongming Rao, Wenliang Zhao, Benlin Liu, Jiwen Lu, Jie Zhou, and Cho-Jui Hsieh.
\newblock Dynamicvit: Efficient vision transformers with dynamic token sparsification.
\newblock \emph{Advances in neural information processing systems}, 34:\penalty0 13937--13949, 2021.

\bibitem[Sheng et~al.(2021)Sheng, Cai, Liu, Deng, Huang, Hua, and Zhao]{ct3d}
Hualian Sheng, Sijia Cai, Yuan Liu, Bing Deng, Jianqiang Huang, Xian-Sheng Hua, and Min-Jian Zhao.
\newblock Improving 3d object detection with channel-wise transformer.
\newblock In \emph{Proceedings of the IEEE/CVF international conference on computer vision}, pages 2743--2752, 2021.

\bibitem[Shi et~al.(2019)Shi, Wang, and Li]{pointrcnn}
Shaoshuai Shi, Xiaogang Wang, and Hongsheng Li.
\newblock Pointrcnn: 3d object proposal generation and detection from point cloud.
\newblock In \emph{Proceedings of the IEEE/CVF conference on computer vision and pattern recognition}, pages 770--779, 2019.

\bibitem[Shi et~al.(2020)Shi, Guo, Jiang, Wang, Shi, Wang, and Li]{pvrcnn}
Shaoshuai Shi, Chaoxu Guo, Li Jiang, Zhe Wang, Jianping Shi, Xiaogang Wang, and Hongsheng Li.
\newblock Pv-rcnn: Point-voxel feature set abstraction for 3d object detection.
\newblock In \emph{Proceedings of the IEEE/CVF conference on computer vision and pattern recognition}, pages 10529--10538, 2020.

\bibitem[Sun et~al.(2020)Sun, Kretzschmar, Dotiwalla, Chouard, Patnaik, Tsui, Guo, Zhou, Chai, Caine, et~al.]{waymo}
Pei Sun, Henrik Kretzschmar, Xerxes Dotiwalla, Aurelien Chouard, Vijaysai Patnaik, Paul Tsui, James Guo, Yin Zhou, Yuning Chai, Benjamin Caine, et~al.
\newblock Scalability in perception for autonomous driving: Waymo open dataset.
\newblock In \emph{Proceedings of the IEEE/CVF conference on computer vision and pattern recognition}, pages 2446--2454, 2020.

\bibitem[Sun et~al.(2022)Sun, Tan, Wang, Liu, Xia, Leng, and Anguelov]{swformer}
Pei Sun, Mingxing Tan, Weiyue Wang, Chenxi Liu, Fei Xia, Zhaoqi Leng, and Dragomir Anguelov.
\newblock Swformer: Sparse window transformer for 3d object detection in point clouds.
\newblock In \emph{European Conference on Computer Vision}, pages 426--442. Springer, 2022.

\bibitem[Vaswani et~al.(2017)Vaswani, Shardlow, Parmar, Uszkoreit, Jones, Gomez, Kaiser, Kattner, Klic, Nandvanshi, Ourselin, Polosukhin, Vinyals, Wainwright, Wu, Yang, Zhai, and Zhou]{attention}
Ashish Vaswani, Michael Shardlow, Niki Parmar, Jakob Uszkoreit, Llion Jones, Aidan~N. Gomez, Łukasz Kaiser, Florian Kattner, B. Klic, Aditya Nandvanshi, Simon Ourselin, Ilya Polosukhin, Oriol Vinyals, M.~J. Wainwright, Z. Wu, Noam Yang, Y. Zhai, and V. Zhou.
\newblock Attention is all you need.
\newblock In \emph{Advances in Neural Information Processing Systems}. Curran Associates, Inc., 2017.

\bibitem[Wang et~al.(2023)Wang, Shi, Shi, Lei, Wang, He, Schiele, and Wang]{dsvt}
Haiyang Wang, Chen Shi, Shaoshuai Shi, Meng Lei, Sen Wang, Di He, Bernt Schiele, and Liwei Wang.
\newblock Dsvt: Dynamic sparse voxel transformer with rotated sets.
\newblock In \emph{Proceedings of the IEEE/CVF Conference on Computer Vision and Pattern Recognition}, pages 13520--13529, 2023.

\bibitem[Wu et~al.(2019)Wu, Qi, and Fuxin]{pointconv}
Wenxuan Wu, Zhongang Qi, and Li Fuxin.
\newblock Pointconv: Deep convolutional networks on 3d point clouds.
\newblock In \emph{Proceedings of the IEEE/CVF Conference on computer vision and pattern recognition}, pages 9621--9630, 2019.

\bibitem[Yan et~al.(2023)Yan, Liu, Sun, Jia, Li, Wang, and Zhang]{cmt}
Junjie Yan, Yingfei Liu, Jianjian Sun, Fan Jia, Shuailin Li, Tiancai Wang, and Xiangyu Zhang.
\newblock Cross modal transformer: Towards fast and robust 3d object detection.
\newblock In \emph{Proceedings of the IEEE/CVF International Conference on Computer Vision}, pages 18268--18278, 2023.

\bibitem[Yan et~al.(2018)Yan, Mao, and Li]{second}
Yan Yan, Yuxing Mao, and Bo Li.
\newblock Second: Sparsely embedded convolutional detection.
\newblock \emph{Sensors}, 18\penalty0 (10):\penalty0 3337, 2018.

\bibitem[Yang et~al.(2022)Yang, Song, Liu, Mao, Li, Li, Sun, Sun, and Zheng]{dbq}
Jinrong Yang, Lin Song, Songtao Liu, Weixin Mao, Zeming Li, Xiaoping Li, Hongbin Sun, Jian Sun, and Nanning Zheng.
\newblock Dbq-ssd: Dynamic ball query for efficient 3d object detection.
\newblock \emph{arXiv preprint arXiv:2207.10909}, 2022.

\bibitem[Yang et~al.(2020)Yang, Sun, Liu, and Jia]{3dssd}
Zetong Yang, Yanan Sun, Shu Liu, and Jiaya Jia.
\newblock 3dssd: Point-based 3d single stage object detector.
\newblock In \emph{Proceedings of the IEEE/CVF conference on computer vision and pattern recognition}, pages 11040--11048, 2020.

\bibitem[Yang et~al.(2021)Yang, Zhou, Chen, and Ngiam]{3dman}
Zetong Yang, Yin Zhou, Zhifeng Chen, and Jiquan Ngiam.
\newblock 3d-man: 3d multi-frame attention network for object detection.
\newblock In \emph{Proceedings of the IEEE/CVF conference on computer vision and pattern recognition}, pages 1863--1872, 2021.

\bibitem[Yin et~al.(2021)Yin, Zhou, and Krahenbuhl]{centerpoint}
Tianwei Yin, Xingyi Zhou, and Philipp Krahenbuhl.
\newblock Center-based 3d object detection and tracking.
\newblock In \emph{Proceedings of the IEEE/CVF conference on computer vision and pattern recognition}, pages 11784--11793, 2021.

\bibitem[Zhang et~al.(2024{\natexlab{a}})Zhang, Liang, Tan, Ye, Zhang, Wang, and Bai]{toc3d}
Dingyuan Zhang, Dingkang Liang, Zichang Tan, Xiaoqing Ye, Cheng Zhang, Jingdong Wang, and Xiang Bai.
\newblock Make your vit-based multi-view 3d detectors faster via token compression.
\newblock \emph{arXiv preprint arXiv:2409.00633}, 2024{\natexlab{a}}.

\bibitem[Zhang et~al.(2024{\natexlab{b}})Zhang, Junnan, Gao, Li, and Hu]{hednet}
Gang Zhang, Chen Junnan, Guohuan Gao, Jianmin Li, and Xiaolin Hu.
\newblock Hednet: A hierarchical encoder-decoder network for 3d object detection in point clouds.
\newblock \emph{Advances in Neural Information Processing Systems}, 36, 2024{\natexlab{b}}.

\bibitem[Zhang et~al.(2022)Zhang, Hu, Xu, Ma, Wan, and Guo]{IA-SSD}
Yifan Zhang, Qingyong Hu, Guoquan Xu, Yanxin Ma, Jianwei Wan, and Yulan Guo.
\newblock Not all points are equal: Learning highly efficient point-based detectors for 3d lidar point clouds.
\newblock In \emph{Proceedings of the IEEE/CVF conference on computer vision and pattern recognition}, pages 18953--18962, 2022.

\bibitem[Zhou and Tuzel(2018)]{voxelnet}
Yin Zhou and Oncel Tuzel.
\newblock Voxelnet: End-to-end learning for point cloud based 3d object detection.
\newblock In \emph{Proceedings of the IEEE conference on computer vision and pattern recognition}, pages 4490--4499, 2018.

\bibitem[Zhou et~al.(2022)Zhou, Zhao, Wang, Wang, and Foroosh]{centerformer}
Zixiang Zhou, Xiangchen Zhao, Yu Wang, Panqu Wang, and Hassan Foroosh.
\newblock Centerformer: Center-based transformer for 3d object detection.
\newblock In \emph{European Conference on Computer Vision}, pages 496--513. Springer, 2022.

\end{thebibliography}
}
\end{document}